# Well-Founded Semantics for Extended Logic Programs with Dynamic Preferences

**Gerhard Brewka**　　　　　　　　　　　　　　　　　　　　　　　BREWKA@KR.TUWIEN.AC.AT
*TU Wien, Abteilung für Wissensbasierte Systeme*
*Treitlstr. 3, 1040 Wien, Austria*

**Abstract**

The paper describes an extension of well-founded semantics for logic programs with two types of negation. In this extension information about preferences between rules can be expressed in the logical language and derived dynamically. This is achieved by using a reserved predicate symbol and a naming technique. Conflicts among rules are resolved whenever possible on the basis of derived preference information. The well-founded conclusions of prioritized logic programs can be computed in polynomial time. A legal reasoning example illustrates the usefulness of the approach.

## 1. Introduction: Why Dynamic Preferences are Needed

Preferences among defaults play a crucial role in nonmonotonic reasoning. One source of preferences that has been studied intensively is specificity (Poole, 1985; Touretzky, 1986; Touretzky, Thomason, & Horty, 1991). In case of a conflict between defaults we tend to prefer the more specific one since this default provides more reliable information. E.g., if we know that students are adults, adults are normally employed, students are normally not employed, we want to conclude "Peter is not employed" from the information that Peter is a student, thus preferring the student default over the conflicting adult default.

Specificity is an important source of preferences, but not the only one, and at least in some applications not necessarily the most important one. In the legal domain it may, for instance, be the case that a more general rule is preferred since it represents federal law as opposed to state law (Prakken, 1993). In these cases preferences may be based on some basic principles regulating how conflicts among rules are to be resolved.

Also in other application domains, like model based diagnosis or configuration, preferences play a fundamental role. Model based diagnosis uses logical descriptions of the normal behaviour of components of a device together with a logical description of the actually observed behaviour. One tries to assume normal behaviour for as many components as possible. A diagnosis corresponds to a set of components for which these normalcy assumptions lead to inconsistency. Very often a large number of possible diagnoses is obtained. In real life some components are less reliable than others. To eliminate less plausible diagnoses one can give the normalcy assumptions for reliable components higher priority.

In configuration tasks it is often impossible to achieve all of the design goals. Often one can distinguish more important goals from less important ones. To construct the best possible configurations goals then have to be represented as defaults with different preferences according to their desirability.

The relevance of preferences is well-recognized in nonmonotonic reasoning, and prioritized versions for most of the nonmonotonic logics have been proposed, e.g., prioritized circumscription (Lifschitz, 1985), hierarchic autoepistemic logic (Konolige, 1988), prioritized default logic (Brewka, 1994a). In these approaches preferences are handled in an "external" manner in the following sense: some ordering among defaults is used to control the generation of the nonmonotonic conclusions. For instance, in the case of prioritized default logic this information is used to control the generation of extensions. However, the preference information itself is not expressed in the logical language. This means that this kind of information has to be fully pre-specified, there is no way of reasoning about (as opposed to reasoning with) preferences. This is in stark contrast to the way people reason and argue with each other. In legal argumentation, for instance, preferences are context-dependent, and the assessment of the preferences among involved conflicting laws is a crucial (if not the most crucial) part of the reasoning.

What we would like to have, therefore, is an approach that allows us to represent preference information *in* the language and derive such information dynamically. In a recent paper (Brewka, 1994b) the author has described a variant of normal default logic in which reasoning about preferences is possible. Although the version of default logic presented in this earlier paper produces reasonable results in most cases, this approach has several drawbacks:

1. The approach is computationally extremely demanding as it involves the construction of the Reiter extensions and an additional compatibility check for each extension guaranteeing that the preference information was taken into account adequately.

2. It may happen that consistent default theories, i.e., theories whose strict part is satisfiable, possess no extensions at all. This is astonishing since in that paper we only dealt with normal defaults. The non-existence of extensions is due to defeasible preference information. It is highly questionable whether such information should be able to destroy all extensions.

3. The earlier paper did not take non-normal defaults into account, it is thus not general enough to cover normal logic programs with negation as failure.

The approach presented in this paper will be based on extended logic programs with two types of negation. This means that in comparison with our earlier proposal we are more restrictive in one respect and more general in another: we are more restrictive since we do not allow arbitrary first order formulas as in normal default logic; we are more general since we admit negation as failure and hence rules which correspond to non-normal defaults in Reiter's logic. We also switch from the extension based semantics of default logic to well-founded semantics (van Gelder, Ross, & Schlipf, 1991; Przymusinski, 1991; Lifschitz, 1996), i.e., to an inherently skeptical approach where the nonmonotonic conclusions are defined directly, not through the notion of extensions. It is well-known that well-founded semantics sometimes looses intuitively expected conclusions. This is also the case in our proposal. However, this is outweighed by a tremendous gain in efficiency: the well-founded conclusions can be computed in polynomial time.

The outline of the rest of the paper is as follows: in Section 2 we first review a definition of well-founded semantics for logic programs with two types of negation which is based on the

20

double application of a certain anti-monotone operator. The definition extends Baral and Subrahmanian's formulation of well-founded semantics for normal logic programs (Baral & Subrahmanian, 1991) and was used by several authors (Baral & Gelfond, 1994; Lifschitz, 1996). We show that this definition suffers from an unnecessary weakness and present a reformulation that leads to better results. Section 3, the main section of the paper, introduces our dynamic treatment of preferences together with several small motivating examples. We show that our conclusions are, in general, a superset of the well-founded conclusions. Section 4 illustrates the expressive power of our approach using a more realistic example from legal reasoning. Section 5 shows that the worst case time complexity for generating well-founded conclusions for prioritized programs is polynomial. Section 6 investigates the relationship to Gelfond and Lifschitz's answer set semantics (Gelfond & Lifschitz, 1990). Section 7 discusses related work and concludes.

## 2. Well-Founded Semantics for Extended Logic Programs

A (propositional) extended logic program consists of rules of the form

$$c \leftarrow a_1, \ldots, a_n, \textit{not } b_1, \ldots, \textit{not } b_m$$

where the $a_i, b_j$ and $c$ are propositional literals, i.e., either propositional atoms or such atoms preceded by the classical negation sign. The symbol *not* denotes negation by failure (weak negation), $\neg$ denotes classical (strong) negation. For convenience we will sometimes use a rule schema to represent a set of propositional rules, namely the set of all ground instances of the schema.

Extended logic programs are very useful for knowledge representation purposes, see for instance (Baral & Gelfond, 1994) for a number of illustrative examples. Two major semantics for extended logic programs have been defined: (1) answer set semantics (Gelfond & Lifschitz, 1990), an extension of stable model semantics, and (2) a version of well-founded semantics (Przymusinski, 1991). The second approach can be viewed as an efficient approximation of the first.

Let us first introduce answer sets. We say a rule $r \in P$ of the form above is defeated by a literal $l$ if $l = b_i$ for some $i \in \{1, \ldots, m\}$. We say $r$ is defeated by a set of literals $X$ if $X$ contains a literal that defeats $r$. Furthermore, we call the rule obtained by deleting weakly negated preconditions from $r$ the monotonic counterpart of $r$ and denote it with $Mon(r)$. We also apply $Mon$ to sets of rules with the obvious meaning.

**Definition 1** *Let $P$ be a logic program, $X$ a set of literals. The $X$-reduct of $P$, denoted $P^X$, is the program obtained from $P$ by*

- *deleting each rule defeated by $X$, and*

- *replacing each remaining rule $r$ with its monotonic counterpart $Mon(r)$.*

**Definition 2** *Let $R$ be a set of rules without negation as failure. $Cn(R)$ denotes the smallest set of literals that is*

1. *closed under $R$, and*

    *2. logically closed, i.e., either consistent or equal to the set of all literals.*

**Definition 3** *Let $P$ be a logic program, $X$ a set of literals. Define an operator $\gamma_P$ as follows:*

$$\gamma_P(X) = Cn(P^X)$$

*$X$ is an answer set of $P$ iff $X = \gamma_P(X)$.*

A literal $l$ is a consequence of a program $P$ under answer set semantics, denoted $l \in Ans(P)$, iff $l$ is contained in all answer sets of $P$.

    The second major semantics for extended logic programs, well-founded semantics, is an inherently skeptical semantics that refrains from drawing conclusions whenever there is a potential conflict. The original formulation of well-founded semantics for general logic programs by Gelder, Ross and Schlipf (1991) is based on a certain partial model. Przymusinski reconstructed this definition in 3-valued logic (Przymusinski, 1990). The formulation using an anti-monotone operator was first given by Baral and Subrahmanian (1991) for general logic programs together with a corresponding definition for default logic. The straightforward extension of this formulation (respectively, the restriction of the default logic definition) to extended logic programs that will be introduced now was used by several authors, e.g. (Baral & Gelfond, 1994; Lifschitz, 1996).[1] Note that in this paper we will only consider the literals that are true in the corresponding 3-valued semantics.

    Like answer set semantics the well-founded semantics for extended logic programs is based on the operator $\gamma_P$ However, the operator is used in a totally different way. Since $\gamma_P$ is anti-monotone the function $\Gamma_P = (\gamma_P)^2$ is monotone. According to the famous Knaster-Tarski theorem (Tarski, 1955) every monotone operator has a least fixpoint. The set of well-founded conclusions of a program $P$, denoted $WFS(P)$, is defined to be this least fixpoint of $\Gamma_P$. The fixpoint can be approached from below by iterating $\Gamma_P$ on the empty set. In case $P$ is finite this iteration is guaranteed to actually reach the fixpoint.

    The intuition behind this use of the operator is as follows: whenever $\gamma_P$ is applied to a set of literals $X$ known to be true it produces the set of all literals that are still potentially derivable. Applying it to such a set of potentially derivable literals it produces a set of literals known to be true, often larger than the original set $X$. Starting with the empty set and iterating until the fixpoint is reached thus produces a set of true literals. It can be shown that every well-founded conclusion is a conclusion under the answer set semantics. Well-founded semantics can thus be viewed as an approximation of answer set semantics.

    Unfortunately it turns out that for many programs the set of well-founded conclusions is extremely small and provides a very poor approximation of answer set semantics. Consider the following program $P_0$ which has also been discussed by Baral and Gelfond (1994):

    1) $b \leftarrow not\,\neg b$
    2) $a \leftarrow not\,\neg a$
    3) $\neg a \leftarrow not\,a$

---

1. Pereira and Alferes (1992) introduced a version of well-founded semantics that adheres to the so-called coherence principle which requires that strong negation implies weak negation. We will show later in Sect. 3 how the coherence principle can be introduced in our approach.

The set of well-founded conclusions is empty since $\gamma_{P_0}(\emptyset)$ equals $Lit$, the set of all literals, and the $Lit$-reduct of $P_0$ contains no rule at all. This is surprising since, intuitively, the conflict between 2) and 3) has nothing to do with $\neg b$ and $b$.

This problem arises whenever the following conditions hold:

1. a complementary pair of literals is provable from the monotonic counterparts of the rules of a program $P$, and

2. there is at least one proof for each of the complementary literals whose rules are not defeated by $Cn(P')$, where $P'$ consists of the "strict" rules in $P$, i.e., those without negation as failure.

In this case well-founded semantics concludes $l$ iff $l \in Cn(P')$. It should be obvious that such a situation is not just a rare limiting case. To the contrary, it can be expected that many commonsense knowledge bases will give rise to such undesired behaviour. For instance, assume a knowledge base contains information that birds normally fly and penguins normally don't, expressed as the set of ground instances of the following rule schemata:

1) $fly(x) \leftarrow not \neg fly(x), bird(x)$
2) $\neg fly(x) \leftarrow not\ fly(x), penguin(x)$

Assume further that the knowledge base contains the information that Tweety is a penguin bird. Now if neither $fly(Tweety)$ nor $\neg fly(Tweety)$ follows from strict rules in the knowledge base we are in the same situation as with $P_0$: well-founded semantics does not draw any "defeasible" conclusion, i.e. a conclusion derived from a rule with weak negation in the body, at all.

We want to show that a minor reformulation of the fixpoint operator can overcome this intolerable weakness and leads to much better results. Consider the following operator

$$\gamma_P^\star(X) = Cl(P^X)$$

where $Cl(R)$ denotes the minimal set of literals closed under the (classical) rules $R$. $Cl(R)$ is thus like $Cn(R)$ without the requirement of logical closedness. Now define

$$\Gamma_P^\star(X) = \gamma_P(\gamma_P^\star(X))$$

Again we iterate on the empty set to obtain the well-founded conclusions of a program $P$ which we will denote $WFS^\star(P)$.

Consider the effects of this modification on our example $P_0$. $\gamma_{P_0}^\star(\emptyset) = \{a, \neg a, b\}$. Rule 1) is contained in the $\{a, \neg a, b\}$-reduct of $P_0$ and thus $\Gamma_{P_0}^\star(\emptyset) = \{b\}$. Since $b$ is also the only literal contained in all answer sets of $P_0$ our approximation actually coincides with answer set semantics in this case.

In the Tweety example both $fly(Tweety)$ and $\neg fly(Tweety)$ are provable from the $\emptyset$-reduct of the knowledge base. However, this has no influence on whether a rule not containing the weak negation of one of these two literals in the body is used to produce $\gamma_P^\star(\emptyset)$ or not. The effect of the conflicting information about Tweety's flying ability is thus kept local and does not have the disastrous consequences it has in the original formulation of well-founded semantics.

It is not difficult to see that the new monotone operator is equivalent to the original one whenever $P$ does not contain negation as failure. In this case the $X$-reduct of $P$, for arbitrary $X$, is equivalent to $P$ and for this reason it does not make any difference whether to use $\gamma_P$ or $\gamma_P^\star$ as the operator to be applied first in the definition of $\Gamma_P$. The same is obviously true for programs without classical negation: for such programs $Cn$ can never produce complementary pairs of literals and for this reason the logical closedness condition is obsolete.

In the general case the new operator produces more conclusions than the original one:

**Proposition 1** *Let $P$ be an extended logic program. For an arbitrary set of literals $X$ we have*
$$\Gamma_P(X) \subseteq \Gamma_P^\star(X).$$

**Proof:** We have $\gamma_P^\star(X) \subseteq \gamma_P(X)$, thus $P^{\gamma_P(X)} \subseteq P^{\gamma_P^\star(X)}$. From this the result follows immediately. $\square$

It remains to be shown that the new operator produces no unwanted results, i.e., that our new semantics can still be viewed as an approximation of answer set semantics.

**Proposition 2** *Let $P$ be an extended logic program. Let $Ans(P)$ be the set of literals contained in all answer sets of $P$. $WFS^\star$ is correct wrt. answer set semantics, i.e., $WFS^\star(P) \subseteq Ans(P)$.*

**Proof:** The proposition is trivially satisfied whenever $P$ has no answer set at all, or when $Lit$ is the single answer set of $P$. So assume $P$ possesses a non-empty set of consistent answer sets, the only remaining possibility according to results in (Gelfond & Lifschitz, 1990).

To show that iterating $\Gamma_P^\star$ on the empty set cannot produce a literal $s \notin Ans(P)$ it suffices to show that $X \subseteq Ans(P)$ implies $\Gamma_P^\star(X) \subseteq Ans(P)$.

Let $A$ be an arbitrary answer set and assume $X \subseteq Ans(P)$. Since $X \subseteq A$ we have $P^A \subseteq P^X$. Since by assumption $A$ is consistent we have $A = Cn(P^A) \subseteq Cl(P^X)$. Therefore $\Gamma_P^\star(X) = Cn(P^{Cl(P^X)}) \subseteq Cn(P^A) = A$. $\square$

For the rest of the paper a minor reformulation turns out to be convenient. Instead of using the monotonic counterparts of undefeated rules we will work with the original rules and extend the definitions of the two operators $Cn$ and $Cl$ accordingly, requiring that weakly negated preconditions be neglected, i.e., for an arbitrary set of rules $P$ with weak negation we define $Cn(P) = Cn(Mon(P))$ and $Cl(P) = Cl(Mon(P))$. We can now equivalently characterize $\gamma_P$ and $\gamma_P^\star$ by the equations

$$\gamma_P(X) = Cn(P_X)$$
$$\gamma_P^\star(X) = Cl(P_X)$$

where $P_X$ denotes the set of rules not defeated by $X$.

Before we turn to the treatment of preferences we give an alternative characterization of $\Gamma_P^\star$ based on the following notion:

**Definition 4** *Let $P$ be a logic program, $X$ a set of literals. A rule $r$ is $X$-safe wrt. $P$ ($r \in SAFE_X(P)$) if $r$ is not defeated by $\gamma_P^\star(X)$ or, equivalently, if $r \in P_{\gamma_P^\star(X)}$.*

With this new notion we can obviously characterize $\Gamma_P^\star$ as follows:

$$\Gamma_P^\star(X) = Cn(P^{\gamma_P^\star(X)}) = Cn(P_{\gamma_P^\star(X)}) = Cn(SAFE_X(P))$$

It is this last formulation that we will modify later. More precisely, the notion of $X$-safeness will be weakened to handle preferences adequately.

## 3. Adding Preferences

In order to handle preferences we need to be able to express preference information explicitly. Since we want to do this *in* the logical language we have to extend the language. We do this in two respects:

1. we use a set of rule names $N$ together with a naming function *name* to be able to refer to particular rules,

2. we use a special (infix) symbol $\prec$ that can take rule names as arguments to represent preferences among rules.

Intuitively, $n_1 \prec n_2$ where $n_1$ and $n_2$ are rule names means the rule with name $n_1$ is preferred over the rule with name $n_2$.[2]

A prioritized logic program is a pair $(R, name)$ where $R$ is a set of rules and *name* a naming function. To make sure that the symbol $\prec$ has its intended meaning, i.e., represents a transitive and anti-symmetric relation, we assume that $R$ contains all ground instances of the schemata

$$N_1 \prec N_3 \leftarrow N_1 \prec N_2, N_2 \prec N_3$$

and

$$\neg(N_2 \prec N_1) \leftarrow N_1 \prec N_2$$

where $N_i$ are parameters for names. Note that in our examples we won't mention these rules explicitly.

The function *name* is a partial injective naming function that assigns a name $n \in N$ to some of the rules in $R$. Note that not all rules do necessarily have a name. The reason is that names will only play a role in conflict resolution among defeasible rules, i.e., rules with weakly negated preconditions. For this reason names for strict rules, i.e., rules in which the symbol *not* does not appear, won't be needed. A technical advantage of leaving some rules unnamed is that the use of rule schemata with parameters for rule names does not necessarily make programs infinite. If we would require names for all rules we would have to use a parameterized name for each schema and thus end up with an infinite set $N$ of names.

In our examples we assume that $N$ is given implicitly. We also define the function *name* implicitly. We write:

$$n_i : c \leftarrow a_1, \ldots, a_n, \textit{not } b_1, \ldots, \textit{not } b_m$$

to express that $name(c \leftarrow a_1, \ldots, a_n, \textit{not } b_1, \ldots, \textit{not } b_m) = n_i$.

---

2. Note that for historical reasons we follow the convention that the minimal rules are the preferred ones.

For convenience we will simply speak of programs instead of prioritized logic programs whenever this does not lead to misunderstandings.

Before introducing new definitions we would like to point out how we want the new explicit preference information to be used. Our approach follows two principles:

1. we want to extend well-founded semantics, i.e. we want that every $WFS^\star$-conclusion remains a conclusion in the prioritized approach,

2. we want to use preferences to solve conflicts whenever this is possible without violating principle 1.

Let us first explain what we mean by conflict here. Rules may be conflicting in several ways. In the simplest case two rules may have complementary literals in their heads. We call this a type-I conflict. Conflicts of this type may render the set of well-founded conclusions inconsistent, but do not necessarily do so. If, for instance, a precondition of one of the rules is not derivable or a rule is defeated the conflict is implicitly resolved. In that case the preference information will simply be neglected. Consider the following program $P_1$:

$n_1 : b \leftarrow not\ c$
$n_2 : \neg b \leftarrow not\ b$
$n_3 : n_2 \prec n_1$

There is a type-I conflict between $n_1$ and $n_2$. Although the explicit preference information gives precedence to $n_2$ we want to apply $n_1$ here to comply with the first of our two principles. Technically, this means that we can apply a preferred rule $r$ only if we are sure that $r$'s application actually leads to a situation where literals defeating $r$ can no longer be derived.

The following two rules exhibit a different type of conflict:

$a \leftarrow not\ b$
$b \leftarrow not\ a$

The heads of these rules are not complementary. However, the application of one rule defeats the other and vice versa. We call this a direct type-II conflict. Of course, in the general case the defeat of the conflicting rule may be indirect, i.e. based on the existence of additional rules. We say $r_1$ and $r_2$ are type-II conflicting wrt. a set of rules $R$ iff

1. $Cl(R)$ neither defeats $r_1$ nor $r_2$,

2. $Cl(R + r_1)$ defeats $r_2$, and

3. $Cl(R + r_2)$ defeats $r_1$

Here $R + r$ abbreviates $R \cup \{r\}$. A direct type-II conflict is thus a type-II conflict wrt. the empty set of rules. The rule sets $R$ that have to be taken into account in our well-founded semantics based approach are subsets of the rules which are undefeated by the set of literals known to be true. Note that the two types of conflict are not disjoint, i.e. two rules may be in conflict of both type-I and type-II. Consider the following program $P_2$, a slight modification of $P_1$:

$$n_1 : b \leftarrow \mathit{not}\ c, \mathit{not}\ \neg b$$
$$n_2 : \neg b \leftarrow \mathit{not}\ b$$
$$n_3 : n_2 \prec n_1$$

Now we have a type-II conflict between $n_1$ and $n_2$ (more precisely, a direct type-II and a type-I conflict) that is not solvable by the implicit mechanisms of well-founded semantics alone. It is this kind of conflict that we try to solve by the explicit preference information. In our example $n_2$ will be used to derive $\neg b$. Note that now the application of $n_2$ defeats $n_1$ and there is no danger that a literal defeating $n_2$ might become derivable later. Generally, a type-II conflict between $r_1$ and $r_2$ (wrt. some undefeated rules of the program) will be solved in favour of the preferred rule, say $r_1$, only if applying $r_1$ excludes any further possibility of deriving an $r_1$-defeating literal.

Note that every type-I conflict can be turned into a direct type-II conflict by a (non-equivalent!) rerepresentation of the rules: if each conflicting rule $r$ is replaced by its seminormal form[3] then all conflicts become type-II conflicts and are thus amenable to conflict resolution through preference information.

After this motivating discussion let us present the new definitions. Our treatment of priorities is based on a weakening of the notion of $X$-safeness. In Sect. 2 we considered a rule $r$ as $X$-safe whenever there is no proof for a literal defeating $r$ from the monotonic counterparts of $X$-undefeated rules. Now in the context of a prioritized logic program we will consider a rule $r$ as $X$-safe if there is no such proof from monotonic counterparts *of a certain subset* of the $X$-undefeated rules. The subset to be used depends on the rule $r$ and consists of those rules that are not "dominated" by $r$. Intuitively, $r'$ is dominated by $r$ iff $r'$ is (1) known to be less preferred than $r$ and (2) defeated when $r$ is applied together with rules that already have been established to be $X$-safe. (2) is necessary to make sure that explicit preference information is used the right way, according to our discussion of $P_1$.

It is obvious that whenever there is no proof for a defeating literal from all $X$-undefeated rules there can be no such proof from a subset of these rules. Rules that were $X$-safe according to our earlier definition thus remain to be $X$-safe. Here are the precise definitions:

**Definition 5** *Let $P = (R, \mathit{name})$ be a prioritized logic program, $X$ a set of literals, $Y$ a set of rules, and $r \in R$. The set of rules dominated by $r$ wrt. $X$ and $Y$, denoted $\mathit{Dom}_{X,Y}(r)$, is the set*

$$\{r' \in R \mid \mathit{name}(r) \prec \mathit{name}(r') \in X \ \mathit{and}\ Cl(Y + r)\ \mathit{defeats}\ r'\}$$

Note that $\mathit{Dom}_{X,Y}(r)$ is monotonic in both $X$ and $Y$. We can now define the $X$-safe rules inductively:

**Definition 6** *Let $P = (R, \mathit{name})$ be a prioritized logic program, $X$ a set of literals. The set of $X$-safe rules of $P$, denoted $\mathit{SAFE}_X^{pr}(P)$, is defined as follows: $\mathit{SAFE}_X^{pr}(P) = \bigcup_{i=0}^{\infty} R_i$, where*

$R_0 = \emptyset$, *and for* $i > 0$,
$R_i = \{r \in R \mid r \ \mathit{not\ defeated\ by}\ Cl(R_X \setminus \mathit{Dom}_{X,R_{i-1}}(r))\}$

---

3. The seminormal form of $c \leftarrow a_1, \ldots, a_n, \mathit{not}\ b_1, \ldots, \mathit{not}\ b_m$ is

$$c \leftarrow a_1, \ldots, a_n, \mathit{not}\ b_1, \ldots, \mathit{not}\ b_m, \mathit{not}\ c'$$

where $c'$ is the complement of $c$. The term seminormal is taken from Reiter (1980).

Note that $X$-safeness is obviously monotonic in $X$. Based on this notion we introduce a new monotonic operator $\Gamma_P^{pr}$:

**Definition 7** *Let $P = (R, name)$ be a prioritized logic program, $X$ a set of literals. The operator $\Gamma_P^{pr}$ is defined as follows:*

$$\Gamma_P^{pr}(X) = Cn(SAFE_X^{pr}(P))$$

As before we define the (prioritized) well-founded conclusions of $P$, denoted $WFS^{pr}(P)$, as the least fixpoint of $\Gamma_P^{pr}$. If a program does not contain preference information at all, i.e., if the symbol $\prec$ does not appear in $R$, the new semantics coincides with $WFS^\star$ since in that case no rule can dominate another rule. In the general case, since the new definition of $X$-safeness is weaker than the one used earlier in Sect. 2 we may have more $X$-safe rules and for this reason obtain more conclusions than via $\Gamma_P^\star$. The following result is thus obvious:

**Proposition 3** *Let $P = (R, name)$ be a prioritized logic program. For every set of literals $X$ we have $\Gamma_R^\star(X) \subseteq \Gamma_P^{pr}(X)$.*

From this and the monotonicity of both operators it follows immediately that $WFS^\star(R) \subseteq WFS^{pr}(P)$.[4]

Well-founded semantics has sometimes been criticized for being too weak and missing intended conclusions. The proposition shows that we can strengthen the obtained results by adding adequate preference information.

As a first simple example let us consider the following program $P_3$:

$n_1 : b \leftarrow not\ c$
$n_2 : c \leftarrow not\ b$
$n_3 : n_2 \prec n_1$

We first apply $\Gamma_{P_3}^{pr}$ to the empty set. Besides the instances of the transitivity and anti-symmetry schema that we implicitly assume only $n_3$ is in $SAFE_\emptyset^{pr}(P_3)$. We thus obtain

$$S_1 = \{n_2 \prec n_1, \neg(n_1 \prec n_2)\}$$

We next apply $\Gamma_{P_3}^{pr}$ to $S_1$. Since $n_2 \prec n_1 \in S_1$ we have $n_1 \in Dom_{S_1, \emptyset}(n_2)$. $n_2 \in SAFE_{S_1}^{pr}(P_3)$ since $Cl(P_{3_{S_1}} \setminus \{n_1\})$ does not defeat $n_2$ and we obtain

$$S_2 = \{n_2 \prec n_1, \neg(n_1 \prec n_2), c\}$$

Further iteration of $\Gamma_{P_3}^{pr}$ yields no new literals, i.e. $S_2$ is the least fixpoint. Note that $c$ is not a conclusion under the original well-founded semantics.

We next show that the programs $P_1$ and $P_2$ discussed earlier are handled as intended. Here is $P_1$:

---

4. Pereira and Alferes (1992) argue that each extension of well-founded semantics to two types of negation should satisfy what they call coherence principle: a weakly negated precondition should be considered satisfied whenever the corresponding strongly negated literal is derived. To model this principle in our approach one would have to weaken the notion of $X$-safeness even further. In the inductive definition, a rule $r$ would have to be considered a member of $R_i$ whenever for each weak precondition *not b* of $r$

- $b \notin Cl(R_X \setminus Dom_{X, R_{i-1}}(r))$, or
- $b' \in X$, where $b' = \neg b$ if $b$ is an atom and $a$ if $b = \neg a$.

$$n_1 : b \leftarrow not\ c$$
$$n_2 : \neg b \leftarrow not\ b$$
$$n_3 : n_2 \prec n_1$$

Since $\gamma^\star_{P_1}(\emptyset)$ does not defeat $n_1$ this rule is safe from the beginning, i.e., $n_1 \in SAFE^{pr}_\emptyset(P_1)$. $\Gamma^{pr}_P(\emptyset)$ yields

$$\{n_2 \prec n_1, \neg(n_1 \prec n_2), b\}$$

which is also the least fixpoint. The explicit preference does not interfere with the implicit one, as intended.

The situation changes in $P_2$ where the first rule in $P_1$ is replaced by

$$n_1 : b \leftarrow not\ c, not \neg b$$

The new rule $n_1$ is not in $SAFE^{pr}_\emptyset(P_2)$ since it is defeated by the consequence of $n_2$ and $n_2$ is not dominated by $n_1$. $\Gamma^{pr}_{P_2}(\emptyset)$ yields

$$S_1 = \{n_2 \prec n_1, \neg(n_1 \prec n_2)\}$$

Now $n_2 \in SAFE^{pr}_{S_1}(P_2)$ since $n_2$ dominates $n_1$ wrt. $S_1$ and the empty set of rules. We thus conclude $\neg b$ as intended. The least fixpoint is

$$S_2 = \{n_2 \prec n_1, \neg(n_1 \prec n_2), b\}$$

In (Brewka, 1994b) we used an example to illustrate the possible non-existence of extensions in our earlier approach. This example involved two normal defaults each of which had the conclusion that the other one is to be preferred. The prioritized logic programming representation of this example is the following:

$$n_1 : n_2 \prec n_1 \leftarrow not \neg(n_2 \prec n_1)$$
$$n_2 : n_1 \prec n_2 \leftarrow not \neg(n_1 \prec n_2)$$

It is straightforward to verify that the set of well-founded conclusions for this example is empty.

## 4. A Legal Reasoning Example

In this section we want to show that the additional expressiveness provided by our approach actually helps representing real world problems. We will use an example first discussed by Gordon (1993, p.7). We somewhat simplified it for our purposes. The same example was also used in (Brewka, 1994b) to illustrate the approach presented there.

Assume a person wants to find out if her security interest in a certain ship is perfected. She currently has possession of the ship. According to the Uniform Commercial Code (UCC, §9-305) a security interest in goods may be perfected by taking possession of the collateral. However, there is a federal law called the Ship Mortgage Act (SMA) according to which a security interest in a ship may only be perfected by filing a financing statement. Such a statement has not been filed. Now the question is whether the UCC or the SMA takes precedence in this case. There are two known legal principles for resolving conflicts of this kind. The principle of *Lex Posterior* gives precedence to newer laws. In our case the UCC is newer than the SMA. On the other hand, the principle of *Lex Superior* gives precedence to laws supported by the higher authority. In our case the SMA has higher authority since it is federal law.

The available information can nicely be represented in our approach. To make the example somewhat shorter we use the notation

$$c \Leftarrow a_1, \ldots, a_n, \mathit{not}\ b_1, \ldots, \mathit{not}\ b_m$$

as an abbreviation for the rule

$$c \leftarrow a_1, \ldots, a_n, \mathit{not}\ b_1, \ldots, \mathit{not}\ b_m, \mathit{not}\ c'$$

where $c'$ is the complement of $c$, i.e. $\neg c$ if $c$ is an atom and $a$ if $c = \neg a$. Such rules thus correspond to semi-normal or, if $m = 0$, normal defaults in Reiter's default logic (Reiter, 1980).

We use the ground instances of the following named rules to represent the relevant article of the UCC, the SMA, Lex Posterior (LP), and Lex Superior (LS). The symbols $d_1$ and $d_2$ are parameters for rule names:

$UCC : \mathit{perfected} \Leftarrow \mathit{possession}$
$SMA : \neg \mathit{perfected} \Leftarrow \mathit{ship}, \neg \mathit{fin\text{-}statement}$
$LP(d_1, d_2) : d_1 \prec d_2 \Leftarrow \mathit{more\text{-}recent}(d_1, d_2)$
$LS(d_1, d_2) : d_1 \prec d_2 \Leftarrow \mathit{fed\text{-}law}(d_1), \mathit{state\text{-}law}(d_2)$

The following facts are known about the case and are represented as rules without body (and without name):

$\mathit{possession}$
$\mathit{ship}$
$\neg \mathit{fin\text{-}statement}$
$\mathit{more\text{-}recent}(UCC, SMA)$
$\mathit{fed\text{-}law}(SMA)$
$\mathit{state\text{-}law}(UCC)$

Let's call the above set of literals $H$. Iterated application of $\Gamma_P^{pr}$ yields the following sequence of literal sets (in each case $S_i = (\Gamma_P^{pr})^i(\emptyset)$):

$$\begin{aligned} S_1 &= H \\ S_2 &= S_1 \end{aligned}$$

The iteration produces no new results besides the facts already contained in the program. The reason is that UCC and SMA block each other, and that no preference information is produced since also the relevant instances of Lex Posterior and Lex Superior block each other. The situation changes if we add information telling us how conflicts between the latter two are to be resolved. Assume we add the following information:[5]

$$LS(SMA, UCC) \prec LP(UCC, SMA)$$

---

5. In realistic settings one would again use a schema here. In order to keep the example simple we use the relevant instance of the schema directly.

Now we obtain the following sequence:

$$\begin{aligned} S_1 &= H \cup \{LS(SMA, UCC) \prec LP(UCC, SMA), \\ &\quad \neg LP(UCC, SMA) \prec LS(SMA, UCC)\} \\ S_2 &= S_1 \cup \{SMA \prec UCC, \neg UCC \prec SMA\} \\ S_3 &= S_2 \cup \{\neg perfected\} \\ S_4 &= S_3 \end{aligned}$$

This example nicely illustrates how in our approach conflict resolution strategies can be specified declaratively, by simply asserting relevant preferences among the involved conflicting rules.

## 5. Complexity

The time complexity of well-founded semantics for a general logic program $P$ is known to be quadratic in the size of $P$, a result attributed to folklore in (Baral & Gelfond, 1994). A proof was given by Witteveen (1991). His analysis is based on Dowling and Gallier's result whereby satisfiability of Horn clauses can be tested in linear time (Dowling & Gallier, 1984). In Dowling and Gallier's approach it is actually a minimal model of a Horn theory that is computed in linear time. Since minimal models of Horn theories are equivalent to closures of rules without negation the result is directly applicable to well-founded semantics for general logic programs. It also applies to well-founded semantics for extended logic programs since for the computation of the least fixed point of $\Gamma_P$ respectively $\Gamma_P^\star$ the complementary literals $l$ and $\neg l$ can be viewed as two distinct atoms.

For the complexity analysis of our prioritized approach let $n$ be the number of rules in a prioritized program $P = (R, name)$. A straightforward implementation would model the application of $\Gamma_P^{pr}$ in an outer loop and the computation of $SAFE_X^{pr}$ in an inner loop. Fortunately, we can combine the two loops into a single loop whose body is executed at most $n$ times. The reason is that $SAFE_X^{pr}$ grows monotonically with $X$ and $\Gamma_P^{pr}$ grows monotonically with $SAFE_X^{pr}$. Here is a nondeterministic algorithm for computing the least fixed point of $\Gamma_P^{pr}$:

> **Procedure** WFS+
> **Input:** A prioritized logic program $P = (R, name)$ with $|R| = n$
> **Output:** the least fixed point of $\Gamma_P^{pr}$
> $S_0 := \emptyset;$
> $R_0 := \emptyset;$
> for $i = 1$ to $n$ do
>
> > if there is a rule $r \in R_{S_{i-1}} \setminus R_{i-1}$ such that
> > $Cl(R_{S_{i-1}} \setminus Dom_{S_{i-1}, R_{i-1}}(r))$ does not defeat $r$
> > then $R_i := R_{i-1} + r; S_i := Cn(R_i)$
> > else return $S_{i-1}$
>
> endfor
> **end** WFS+

In each step $S_i$ and $R_i$ denote the well-founded conclusions, respectively safe rules established so far. The body of the for-loop is executed at most $n$ times and there are at most $n$ rules that have to be checked for satisfaction of the if-condition. The if-condition itself can, according to the results of Dowling and Gallier, be checked in linear time: we need to establish $Dom_{S_{i-1}, R_{i-1}}(r)$ which involves the computation of a minimal model of the monotonic counterparts of $R_{i-1} + r$. We then have to eliminate the rules dominated by $r$ form $R_{S_{i-1}}$ and compute another minimal model to see whether $r$ is defeated.

More precisely, Dowling and Gallier show that the needed time is linear in the number of propositional constants. This number may be greater than $n$ in principle. However, since literals that do not appear in the head of a rule must be false in the minimal model we can eliminate them accordingly and work with a set of rules that has at most $n$ literals. This leads to an overall time complexity of $O(n^3)$.

It should be mentioned, however, that due to the use of rule schemata for transitivity and anti-symmetry of $\prec$ prioritized programs can be considerably larger than corresponding unprioritized programs. The transitivity schema, for instance, has $|N|^3$ instances. An implementation should, therefore, be based on an approach where instances are only generated when actually needed, or on other built in techniques that handle transitivity and anti-symmetry. Such techniques are beyond the scope of this paper.

## 6. Relation to Answer Sets

In this section we will investigate the relation of our modification of well-founded semantics to answer set semantics (Gelfond & Lifschitz, 1990). Since our approach handles an extended language in which certain symbols are given a particular pre-defined meaning a thorough investigation of this relationship is only possible after a corresponding extension of answer set semantics to prioritized logic programs has been defined. We are not planning to introduce and defend such an extension in this paper. Nevertheless, we can give some preliminary results here. More precisely, we will show that the conclusions produced in our proposal are correct wrt. a particular subclass of answer sets, the so-called priority-preserving answer sets.

**Definition 8** *Let $R$ be a logic program, $A$ an answer set of $R$, and let*

$$r = c \leftarrow a_1, \ldots, a_n, \text{not } b_1, \ldots, \text{not } b_m$$

*be a rule in $R$. We say $r$ is rebutted in $A$ ($r \in re_R(A)$) iff $\{a_1, \ldots, a_n\} \subseteq A$ and $r$ is defeated in $A$.*

**Definition 9** *Let $P = (R, name)$ be a prioritized logic program, $A$ an answer set of $R$. $A$ is called priority preserving iff for every $r \in re_R(A)$ the set*

$$Cl(R_A \setminus Dom_{A, R_A}(r))$$

*defeats $r$.*

The intuition behind the definition is the following: whenever a rule $r$ is rebutted in an answer set $A$ but its rebuttal is solely based on rules dominated by $r$ with respect to $A$ and the rules not defeated by $A$ we consider this as a violation of the available preference information and "reject" the answer set.

We can now show correctness of our approach wrt. priority preserving answer sets.

**Proposition 4** *Let $P = (R, name)$ be a prioritized logic program. $l \in WFS^{pr}(P)$ implies $l$ is contained in all priority preserving answer sets of $R$.*

**Proof:** The proof is similar to the correctness proof of $WFS^\star$ wrt. answer set semantics (Proposition 2). Again the proposition is trivially satisfied whenever there is no priority preserving answer set, or $Lit$ is the single priority preserving answer set. We may therefore assume that every priority preserving answer set of $R$ is consistent.

In the inductive step we show that for an arbitrary priority preserving answer set $A$ a rule $r$ is not defeated in $A$ whenever $r \in SAFE_X^{pr}(P)$, given that $X$ is a set of literals true in $A$. From this it follows that $Cn(SAFE_X^{pr}(P))$ contains only literals true in all priority preserving answer sets.

Let $R_i$ be defined as in Def. 6 (the inductive definition of $X$-safeness) and assume it is already known that the rules in $R_{i-1}$ are not defeated in $A$. By definition $r = c \leftarrow a_1, \ldots, a_n, not\ b_1, \ldots, not\ b_m \in R_i$ iff $r$ is not defeated by $Cl(R_X \setminus Dom_{X, R_{i-1}}(r))$. We distinguish two cases:

Case 1: $a_1, \ldots, a_n \in A$: Since $X \subseteq A$ and $Dom$ is monotonic in both indices we have $Cl(R_A \setminus Dom_{A, R_A}(r)) \subseteq Cl(R_X \setminus Dom_{X, R_{i-1}}(r))$. Therefore $r$ cannot be defeated in $A$ since $A$ is priority preserving.

Case 2: $a_1, \ldots, a_n \notin A$: Since the prerequisites of $r$ cannot be derived from $R_A$ the set $Dom_{A, R_A}(r)$ contains only rules defeated by $Cl(R_A)$ alone. Since $A$ is an answer set these rules can't be contained in $R_A$. Therefore $Cl(R_A) = Cl(R_A \setminus Dom_{A, R_A}(r))$ and thus $Cl(R_A) \subseteq Cl(R_X \setminus Dom_{X, R_{i-1}}(r))$. Since by assumption $A$ is consistent we also have $Cl(R_A) = Cn(R_A)$ and therefore $r$ cannot be defeated in $A$. □

We have seen that our approach is guaranteed to produce only conclusions contained in all priority preserving answer sets. We can also ask the opposite question: given a particular answer set $A$, is it always possible to obtain $A$ (or, more precisely, a superset of $A$ containing additional preference information) through prioritized well-founded semantics by adding adequate preference information?

The answer to this question is no. The reason is that for sake of tractability we always consider single rules when determining $X$-safeness in our approach. Here is an example:

$$n_1 : b \leftarrow not\ a$$
$$n_2 : c \leftarrow not\ b$$
$$n_3 : d \leftarrow not\ c$$
$$n_4 : a \leftarrow not\ d$$

This program has two answer sets $S_1 = \{b, d\}$ and $S_2 = \{c, a\}$. Consider $S_1$. Even if we add the preference information that both $n_1$ and $n_3$ are preferred to each of $n_2$ and $n_4$ we are unable to derive $b$ and $d$. For instance, $n_1$ is not $X$-safe because its head does not defeat $n_4$.

In order to derive $S_1$ it would be necessary to take the possibility of sets of rules (here $n_1$ and $n_3$) defeating less preferred sets of rules (here $n_2$ and $n_4$) into account. Although this is possible in principle it would clearly lead to intractability since in the worst case an exponential number of subsets of rules would have to be checked. Giving up tractability seems too high a price for what is gained and we stick to our more cautious approach for this reason.

## 7. Related Work and Conclusions

Several approaches treating preferences in the context of logic programming have been described in the literature. We will now discuss how they relate to our proposal.

Kowalski and Sadri (1991) proposed to consider rules with negation in the head as exceptions to more general rules and to give them higher priority. Technically, this is achieved by a redefinition of answer sets. It turns out that the original answer sets remain answer sets according to the new definition whenever they are consistent. The main achievement is that programs whose single answer set is inconsistent become consistent in the new semantics. The approach can hardly be viewed as a satisfactory treatment of preferences for several reasons:

1. preferences are implicit and highly restricted; the asymmetric treatment of positive and negative information seems unjustified,

2. it is difficult to see how, for instance, exceptions of exceptions can be represented,

3. fewer conclusions are obtained than in the original answer set semantics, contrary to what one would expect when preferences are taken into account.

It is, therefore, more reasonable to view Kowalski and Sadri's approach as a contribution to inconsistency handling rather than preference handling.

An approach that is closer in spirit to ours is ordered logic programming (Buccafurri, Leone, & Rullo, 1996). An ordered logic program is a set of components forming an inheritance hierarchy. Each component consists of a set of rules. The inheritance hierarchy is used to settle conflicts among rules: rules lower in the hierarchy have preference over those higher up in the hierarchy since the former are considered more specific. A notion of a stable model for ordered logic programs can be defined (see Buccafurri et al., 1996, for the details).

There are two main differences between ordered logic programs and our extension of well-founded semantics:

1. ordered logic programs use only one kind of negation, the distinction between negation as failure and classical negation is not expressible in the language,

2. the preferences of ordered logic programs are predefined through the inheritance hierarchy, there is no way of deriving context-dependent preferences dynamically.

Finally we would like to mention an approach recently presented by Prakken and Sartor (1995). They extend Dung's argument system style reconstruction of logic programming (Dung, 1993) with a preference handling method that is very close to ours. This is not astonishing since, as the authors point out, their approach is based on "unpublished ideas of Gerhard Brewka". In fact, it was a preliminary version of this paper that led to their formulation.

We presented in this paper an extension of logic programs with two types of negation where preference information among rules can be expressed in the logical language. This extension is very useful for practical applications, as was demonstrated using an example from legal reasoning. The main advantage of our approach is that also this type of information is context-dependent and can be reasoned upon and derived dynamically.

From well-founded semantics we inherit some drawbacks and advantages. Sometimes reasonable conclusions are not obtained. On the other hand, the addition of preference information can make the set of conclusions considerably larger, as we have shown. Moreover, - and this certainly is the greatest advantage of well-founded semantics and our proposed extension - reasoning can be done in polynomial time.

The simple and natural representation of the legal example discussed in Sect. 4 seems to indicate that our generalization of well-founded semantics may provide a new attractive compromise between expressiveness and efficiency with a number of interesting potential applications.

## Acknowledgements

I would like to thank Franz Baader, Jürgen Dix, Tom Gordon, Henry Prakken, Cees Witteveen, and two anonymous referees for interesting comments helping to improve the quality of this paper.